\documentclass[letterpaper, 10pt, conference]{ieeeconf}  

\IEEEoverridecommandlockouts                              

\usepackage{amsmath,amsfonts}
\usepackage{algorithmic}
\usepackage{algorithm}
\usepackage{array}
\usepackage[caption=false,font=normalsize,labelfont=sf,textfont=sf]{subfig}
\usepackage{textcomp}
\usepackage{stfloats}
\usepackage{url}
\usepackage{verbatim}
\usepackage{graphicx}
\usepackage{cite}
\usepackage{setspace}
\usepackage{threeparttable}
\usepackage{booktabs}

\usepackage{multirow}
\usepackage{xcolor}

\usepackage{geometry}
\title{\LARGE \bf
PGD-VIO: An Accurate Plane-Aided Visual-Inertial Odometry with Graph-Based Drift Suppression
}

\author{Yidi Zhang$^{1, 2}$, Fulin Tang$^{2*}$, 
Zewen Xu$^{2}$, 
Yihong Wu$^{2, 1*}$ and Pengju Ma$^{3}$
\thanks{$^{*}$This work was supported by the National Natural Science Foundation of China under Grant No. 62202468 and a SINOPEC Research Project. The corresponding authors are Fulin Tang and Yihong Wu.
E-mail: {\{\tt\small fulin.tang, yhwu\}@nlpr.ac.cn}}
\thanks{$^{1}$School of Artificial Intelligence, University of Chinese Academy of Sciences, Beijing, China.}%
\thanks{$^{2}$State Key Laboratory of Multimodal Artificial Intelligence Systems, Institute of Automation, Chinese Academy of Science, Beijing, China.}
\thanks{$^{3}$Sinopec Shengli Oilfield, Shandong, China.}%
}

\begin{document}

\maketitle
\thispagestyle{empty}
\pagestyle{empty}

\begin{abstract}

Generally, high-level features provide more geometrical information compared to point features, which can be exploited to further constrain motions. 
Planes are commonplace in man-made environments, offering an active means to reduce drift, due to their extensive spatial and temporal observability.
To make full use of planar information, we propose a novel visual-inertial odometry (VIO) using an RGB-D camera and an inertial measurement unit (IMU), effectively integrating point and plane features in an extended Kalman filter (EKF) framework.
Depth information of point features is leveraged to improve the accuracy of point triangulation, while plane features serve as direct observations added into the state vector. 
Notably, to benefit long-term navigation, a novel graph-based drift detection strategy is proposed to search overlapping and identical structures in the plane map so that the cumulative drift is suppressed subsequently.
The experimental results on two public datasets demonstrate that our system outperforms state-of-the-art methods in localization accuracy and meanwhile generates a compact and consistent plane map, 
free of expensive global bundle adjustment and loop closing techniques.

\end{abstract}


\section{INTRODUCTION}
Visual-inertial odometry (VIO) and simultaneous localization and mapping (SLAM) are key problems in the field of mobile robotics \cite{wu2018image}.
Most existing VIO/SLAM systems rely on sparse point features for the sake of efficiency and robustness \cite{geneva2020openvins, 2021orbslam3}.
RGB-D cameras simplify the tasks of triangulating point features and extracting high-level features.
Compared with point features, plane features can provide complementary information 
to boost the performance,
especially when points degenerate in challenging scenes \cite{cho2021sp}.
Moreover, plane features exist prevalently in man-made environments and are more interpretable and usable in providing a structural representation. 
Therefore, combining point and plane features has been investigated in many studies \cite{yang2019tightly}.


\begin{figure}[thpb]
  \centering
   \includegraphics[width=0.5\textwidth]{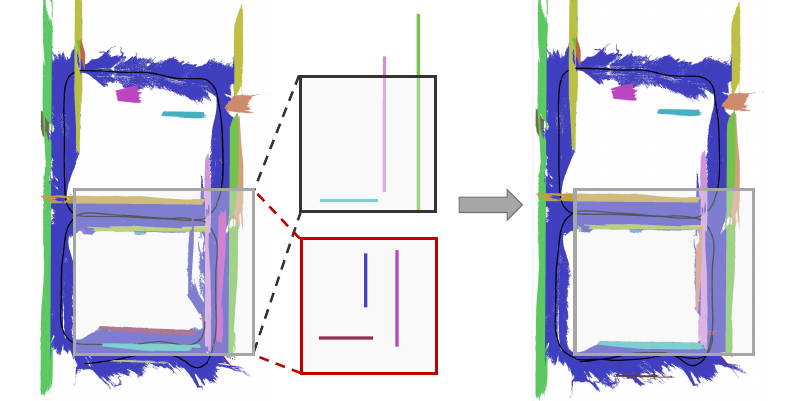}
  \caption{Illustration of the proposed PGD-VIO on the CID-SIMS sequence \textit{Floor3\_1}.
  Attributed to the drift suppression strategy, the system can detect overlapping and identical configurations in the plane map and align them to cope with cumulative errors, resulting in an accurate trajectory and a more consistent plane map.
  \label{fig:result}}
\end{figure}

One of the critical problems for investigating planes as landmarks in RGB-D VIO/SLAM systems is the data association.
Different from point features that are tracked on 2D images, planes are typically associated according to their parameters in a unified coordinate system.
Broadly, two planes are considered as a matching pair when their angle and separation are within the defined thresholds.
In some researches, planar covariance and the Mahalanobis distance are also employed \cite{yang2019tightly}.
As these ideas firmly depend on initial poses, planes cannot be associated successfully once drift occurs, thus submerging their value in providing long-term constraints.
To tackle this issue and fully exploit the longstanding planes, we propose a novel plane-aided RGB-D VIO system with a graph-based drift suppression strategy.
The key idea is to understand the structural regularity of a given scene and perform drift detection by identifying duplicate planar structures in the map.
In other words, if a set of planes overlaps another to some extent and their spatial configurations are similar, the map becomes inconsistent, indicating potential drift.
Once drift is detected, we attempt to suppress it and correct the poses accordingly.
As shown in Fig. \ref{fig:result},
our system can cope with large drift and robustly 
associate repetitive planes to improve the localization performance as well as the map consistency in a corridor environment. 
The main contributions of this work are summarized as follows:
\begin{itemize}
\item
We construct an RGB-D VIO system, called PGD-VIO, within 
an extended Kalman filter (EKF)
framework
and derive how to update the state properly using plane and point features.
\item 
We present a novel graph-based strategy for drift detection using planar structures.
Then, cumulative errors are suppressed through a de-drift update.
By investigating similarities between plane patches, our method can detect repetitive structures in the global map and correct their drift, thereby better constraining the motions.

\item 
We validate the proposed system extensively on two public datasets, demonstrating that our system performs well in localization and builds a consistent plane map by fusing depth and planar properties.

\end{itemize}


\begin{figure*}[thpb]
  \centering
\includegraphics[width=0.75\textwidth]{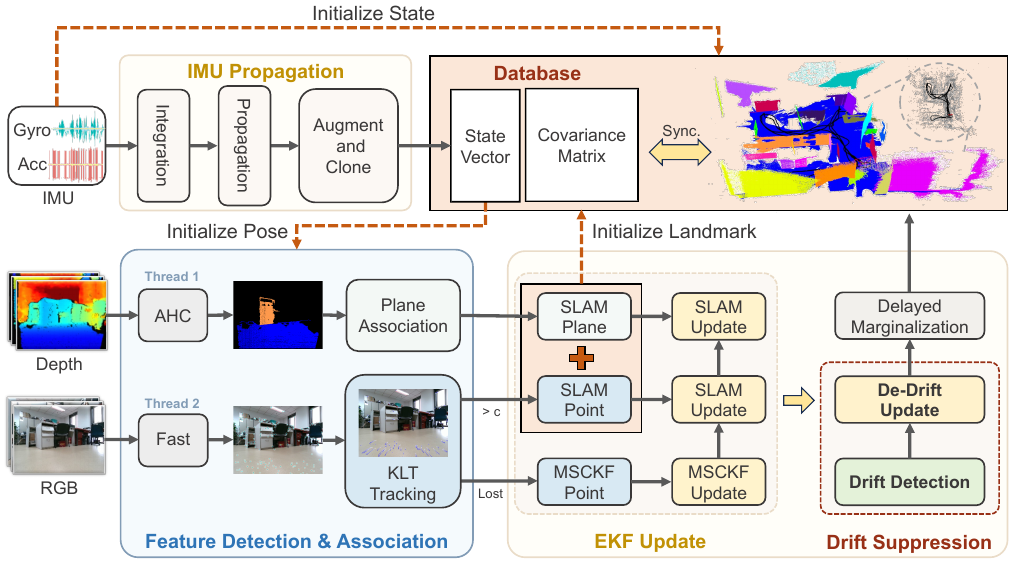}
  \caption{Overview of the proposed PGD-VIO system.}
  \label{fig:pipeline}
\end{figure*}

\section{RELATED WORK}
Nowadays, several methods have combined RGB-D and inertial measurements for navigation
\cite{2021orbslam3, shan2019rgbd, chen2022vip}.
Planes are predominant primitives in man-made environments, which contain prolific geometrical information and structural regularities for gaining in improvement. 
As their parameters can be computed from RGB-D cameras \cite{
hsiao2018dense, yang2019tightly} or 3D LiDARs \cite{geneva2018lips}, multiple researchers leverage planes as direct observations.
Additionally, some studies enforce dependencies on coplanar regularities.
\cite{yang2019tightly} introduces plane measurements adopting closest point (CP) for parameterization and distinguishes planar point features from non-planar point features to permit point-on-plane constraints.
VIP-SLAM \cite{chen2022vip} exploits point-to-plane and homography constraints in a tightly coupled system, which significantly reduces the complexity of bundle adjustment.
\cite{chen2023monocular} is a monocular VIO system 
regularized by point-on-plane constraints within a lightweight multi-state constraint Kalman filter (MSCKF).

Recently, cross-plane constraints have received attention to support camera pose estimation in structured environments.
For example, DPI-SLAM \cite{hsiao2018dense} forces orthogonality and parallelism constraints on nearby planes in global graph optimization.
Besides, some approaches have been developed 
in conjunction with Manhattan world (MW) \cite{kim2018linear} 
or Atlanta world \cite{joo2020linear} assumption.
By recognizing the dominant directions of frames and estimating drift-free rotations followed by the translation-only BA, rotations and translations are decoupled in 
\cite{li2021rgb}.
Based on the understanding of environmental assumptions, \cite{joo2021linear}
detects 
planes that are aligned with the dominant directions to estimate drift-free rotations and update translations and 1D representations of the structure-aware planes in a linear EKF framework.

In light of these attempts, geometric structures are taken into account for plane association.
\cite{2016SceneStructureRegistration} constructs graphs using plane patches and relies on an interpretation tree to search matches
for real-time place recognition.
Such correspondences with prior maps help with error correction for VIO/SLAM systems.
LiPMatch \cite{jiang2020lipmatch} detects loop closures by evaluating plane similarities of two keyframes with a graph matching method in a LiDAR SLAM.
Combining objects and planes, 
\cite{deng2022object} generates a semantic topological graph, in which node descriptors are extracted based on the graph propagation theory. Then, a relocalization system
is developed for pose optimization.
Moreover, \cite{shaheer2023graph} proposes a novel graph-to-graph matching method to relate SLAM maps with architectural plans and achieve global robot localization.
Similarly, PPM-VIO \cite{hu2023efficient} is a filter-based VIO system that exploits a prior point-plane map to correct drift in the local pose estimates.

Inspired by the above researches, to further exploit the structure of planes, 
we propose a novel RGB-D VIO system, incorporating point and plane measurements,
along with a graph-based drift suppression strategy, which can significantly improve performance in long-term navigation.
As distinct from those relying on prior maps to optimize poses, our method detects drift from the incremental plane map and updates the system state in a filtering framework.

%




\section{PROPOSED SYSTEM}
Built upon the EKF framework, our system incorporates depth information into point features and 
utilizes plane measurements in the EKF update with camera-IMU calibration. 
Moreover, to further exploit the structure of scenes, we investigate a graph-based method to detect drift from the global plane map and suppress the errors in long-term localization.
Fig. \ref{fig:pipeline} shows the overview of the proposed system.
PGD-VIO contains four procedures: IMU integration and propagation (Sec. \ref{secB}), feature detection and association (Sec. \ref{secC}), EKF state update (Sec. \ref{secD} and \ref{secE}), and drift suppression (Sec. \ref{secF}).
Given an input RGB-D and inertial sequence, we first apply IMU measurements to propagate the system state and the covariance.
Then, we detect and associate both points and planes in parallel and update them during the EKF update process.
If needed, drift is detected based on a novel graph matching strategy and suppressed with a de-drift update. After that, small and short-tracked planes are delayed marginalized when they are lost for a period of time.

\subsection{State Vector}

At time $t_k$, the system state is defined as follows:
\begin{equation}
\mathbf{x}_{k} = 
\begin{bmatrix}
\mathbf{x}_{I_k}^{\mathsf{T}} &
\mathbf{x}_{calib}^{\mathsf{T}} & 
\mathbf{x}_C^{\mathsf{T}} &
\mathbf{x}_P^{\mathsf{T}} &
\mathbf{x}_{\Pi}^{\mathsf{T}}
\end{bmatrix}^{\mathsf{T}}
\end{equation}
\begin{equation}
\mathbf{x}_{I_k} = 
\begin{bmatrix}
{}^{I_{k}}_G\bar{\mathbf{q}}^{\mathsf{T}}&
{}^G{\mathbf{p}}_{I_{k}}^{\mathsf{T}}&
{}^G{\mathbf{v}}_{I_{k}}^{\mathsf{T}}&
{}^{I_{k}}{\mathbf{b}}_{g}^{\mathsf{T}}&
{}^{I_{k}}{\mathbf{b}}_{a}^{\mathsf{T}}
\end{bmatrix}^{\mathsf{T}}
\end{equation}
\begin{equation}
\mathbf{x}_{calib} = 
\begin{bmatrix}
{}^C_I\bar{\mathbf{q}}^{\mathsf{T}}& 
{}^C{\mathbf{p}}_I^{\mathsf{T}}&
{}_I^Ct &
\mathbf{\lambda}_C^\mathsf{T}
\end{bmatrix}^{\mathsf{T}}
\end{equation}
\begin{equation}
\mathbf{x}_{C} = 
\begin{bmatrix}
^{I_{k}}_G\bar{\mathbf{q}}^{\mathsf{T}} &
^G\mathbf{p}_{I_k}^{\mathsf{T}} & 
... & 
^{I_{k-c}}_G\bar{\mathbf{q}}^{\mathsf{T}} &
^G\mathbf{p}_{I_{k-c}}^{\mathsf{T}}
\end{bmatrix}^{\mathsf{T}}
\end{equation}
\begin{equation}
\mathbf{x}_{P} = 
\begin{bmatrix}
^G\mathbf{f}_1^{\mathsf{T}} \
... \
^G\mathbf{f}_h^{\mathsf{T}}
\end{bmatrix}^{\mathsf{T}} 
, \
\mathbf{x}_{\Pi} = 
\begin{bmatrix}
^G\mathbf{\Pi}_1^{\mathsf{T}} \
... \
^G\mathbf{\Pi}_n^{\mathsf{T}} 
\end{bmatrix}^{\mathsf{T}}.
\end{equation}
For current IMU state $\mathbf{x}_{I_k}$ and historical IMU pose clones
$\mathbf{x}_{C}$,
${}^I_G\bar{\mathbf{q}}$ 
is the unit quaternion representing the rotation from the global frame $\{G\}$ to the IMU frame $\{I\}$, ${}^G{\mathbf{p}}_I$ and $^G{\mathbf{v}}_I$ are the position and velocity of IMU with respect to $\{G\}$, and ${}^I{\mathbf{b}}_{g}$ and ${}^I{\mathbf{b}}_{a}$ are the gyroscope and accelerometer biases, respectively.
$\mathbf{x}_{calib}$ is calibration parameters consisting of camera-IMU rigid transformation 
$\{{}^C_I\bar{\mathbf{q}},\ {}^C{\mathbf{p}}_I\}$, time offset ${}^C_It$ and camera intrinsic parameters $\mathbf{\lambda}_C$.
$\mathbf{x}_{P}$ and $\mathbf{x}_{\Pi}$ are point and plane features in $\{G\}$.
To simplify subsequent expressions, we clarify that throughout the paper, ${}_A^B\mathbf{R}$ is the rotation matrix from frame $\{A\}$ to frame $\{B\}$ and ${}^B\mathbf{p}_A$ is the position of frame $\{A\}$ in frame $\{B\}$.

\subsection{IMU Propagation}\label{secB}
The system state evolves from time $t_k$ to $t_{k+1}$ through IMU integration and forward propagation.
Details about the generic nonlinear IMU kinematics and the evolution process can be found in \cite{mourikis2007multi, geneva2020openvins}.

\subsection{Feature Detection and Association}\label{secC}
In our case, FAST corners
are extracted on color images as keypoints and sparse KLT optical flow is employed to track them between frames.
Meanwhile, planes are detected from depth maps using agglomerative hierarchical clustering (AHC) algorithm \cite{feng2014fast}
and then associated with map planes by comparing their distances and normal vector angles in \{G\}.

\subsection{Point Feature Update}\label{secD}
In the first step we need to obtain initial 3D position estimations of points. 
To this end, we take all the camera poses provided by the IMU propagation to be of known quantity and fuse depth information into the 3D Cartesian Triangulation.
In particular, when a point ${}^G\mathbf{f}_i$ is observed by a camera $C_m$, we have an observation 
\begin{equation}
\begin{aligned}
    {}^{C_m}\mathbf{f}_i = 
    {}^{C_m}{z_f}
    {}^{C_m}\mathbf{b}_f,
\end{aligned}
\end{equation}
where
${}^{C_m}{z_f}$ represents the depth of this point from the image plane and ${}^{C_m}\mathbf{b}_f$ is the bearing vector.
With the knowledge of ${}^{C_m}{z_f}$, we can directly transform the 3D observation ${}^{C_m}{\mathbf{f}}_{i}$ to $\{G\}$ via
\begin{equation}
\begin{aligned}
    {}^G\mathbf{f}_i &= 
    {}^{C_m}_{G}\mathbf{R}^{\mathsf{T}}
    {}^{C_m}{\mathbf{f}}_{i} + 
    {}^G\mathbf{p}_{C_m}.
\end{aligned}
\end{equation}
Otherwise, if ${}^{C_m}{z_f}$ is not available due to noise or exceeding the range, ${}^G\mathbf{f}_i$ can be written as
\begin{equation}\label{eq9}
\begin{aligned}
    {}^G\mathbf{f}_i
    &= 
    {}^{C_m}{z_f} ^G\mathbf{b}_f + {}^G\mathbf{p}_{C_m}.
\end{aligned}
\end{equation}

By defining vectors
orthogonal to ${}^G\mathbf{b}_f$ in $\mathbf{N}_m$ ($\mathbf{N}_m  {}^G\mathbf{b}_f = \mathbf{0}_{3 \times 3}$)
and substituting it to (\ref{eq9}), we can obtain:
\begin{equation}
\begin{aligned}
\mathbf{N}_m {}^G\mathbf{f}_i &= 
\mathbf{N}_m {}^G\mathbf{p}_{C_m}.
\end{aligned}
\end{equation}
After stacking all the hybrid points:
\begin{equation}
\begin{aligned} 
\underbrace{ \begin{bmatrix} \vdots \\ \mathbf{N}_{m_1} \\ \vdots  \\ \mathbf{I}_{3 \times 3} \\ \vdots \end{bmatrix} }_{\mathbf{A}}
{}^G\mathbf{f}_i & = 
\underbrace{ \begin{bmatrix} \vdots \\ \mathbf{N}_{m_1} {}^G\mathbf{p}_{C_{m_1}} \\ \vdots
\\ {}^{C_{m_2}}_G\mathbf{R}^{\mathsf{T}}{{}^{C_{m_2}}\mathbf{f}_i} + {}^G\mathbf{p}_{C_{m_2}} \\ \vdots \end{bmatrix} }_{\mathbf{b}},
\end{aligned}
\end{equation}
the position ${}^G\mathbf{f}_i$ can be calculated by solving the linear system $ \mathbf{A}^\top\mathbf{A}~ {}^G\mathbf{f}_i = \mathbf{A}^\top\mathbf{b}$.

Then, a point feature ${}^G\mathbf{f}_i$ is updated using the following measurement function:
\begin{equation}
\begin{aligned}
\mathbf{z}_b = h(\mathbf{x}_k) + \mathbf{n}_b,
\end{aligned}
\end{equation}
where $h(\cdot)$ projects ${}^G\mathbf{f}_i$ onto an observed image $C_m$ with the state $\mathbf{x}_{T_m}$, including the observing pose $\mathbf{x}_{C_m}$ and the calibration parameters $\mathbf{x}_{calib}$, and $\mathbf{n}_b \sim \mathcal{N}(\mathbf{0}_{2 \times 2},\mathbf{I}_{2 \times 2})$ denotes the measurement noise.

Linearizing the equations yields the following system:
\begin{equation}\label{eqpoint}
\tilde{\mathbf{z}}_b
= \mathbf{H}_{T_b}
\tilde{\mathbf{x}}_{T_m}
+ \mathbf{H}_{f_b} {}^G\tilde{\mathbf{f}}_i
+ \mathbf{n}_b,
\end{equation}
where $\tilde{\mathbf{z}}_b$ is the projected 2D residual, 
$\mathbf{n}_b$ is the white Gaussian noises,
$\mathbf{H}_{T_b}$ and $\mathbf{H}_{f_b}$ are the measurement Jacobians in respect to the current state $\mathbf{x}_{T_m}$ and the 3D point feature ${}^G\mathbf{f}_i$, respectively.

After stacking all the measurements from different timesteps,
we perform an EKF update for point features, which are divided into SLAM features and MSCKF features based on their track lengths.
Since only SLAM features are in the state vector, SLAM points are updated using standard EKF {while feature dependency will be removed from (\ref{eqpoint}) through nullspace projection for MSCKF points.}
For more details please refer to \cite{geneva2020openvins}.

\begin{figure*}[thpb]
  \centering \includegraphics[width=1.0\textwidth]{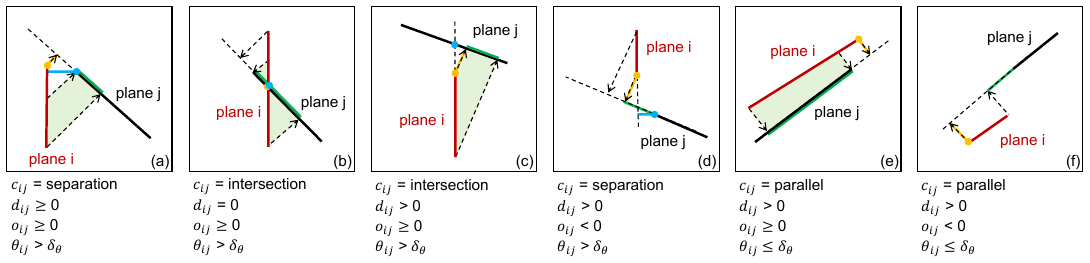}
  \caption{Distinctive relative positions of two plane patches, viewed from a common perpendicular direction to their normal vectors.
  In each figure, the yellow dot is the closest point on plane $i$ to plane $j$ and the corresponding line depicts the distance $d_{ij}$, the blue dot and the blue line measure the distance $d_{ji}$ from plane $j$ to plane $i$ oppositely, and the green line marks their overlapping region, which is negative implying the parallel distance in (d) and (f).
  \label{fig:relation}}
\end{figure*}

\subsection{Plane Feature Update}\label{secE}
Observations of environmental planes can be obtained directly from depth images provided by the RGB-D camera.
Here, closet point (CP) \cite{geneva2018lips} is adopted to represent a plane feature:
\begin{equation}
\begin{aligned}
{}^G\mathbf{\Pi} = {}^G\mathbf{n} {}^Gd, \
\begin{bmatrix}
{}^G\mathbf{n}\\
{}^Gd
\end{bmatrix}
=
\begin{bmatrix}
{}^G\mathbf{\Pi}/||{}^G\mathbf{\Pi}|| \\
||{}^G\mathbf{\Pi}||
\end{bmatrix},
\end{aligned}
\end{equation}
where ${}^G\mathbf{n}$ and ${}^Gd$ are the unit normal vector and the distance scalar of the plane, respectively. 
To fit a plane, we minimize point-to-plane distances by solving a maximum likelihood estimation (MLE) problem with RANSAC on the basic of AHC \cite{feng2014fast}.
Once a plane ${}^C\mathbf{\Pi}$ is observed by a camera $C_m$,
we recover the initial guess for it.
Afterwards, a plane feature ${}^G\mathbf{\Pi}$ can be updated using the following measurement function:
\begin{equation}
\begin{aligned}
    {}^{C_m}\mathbf{\Pi} 
    =& 
    ({}_G^{C_m}\mathbf{R} {}^G\mathbf{n})({}^Gd - {}^G\mathbf{p}_{C_m}^{\mathsf{T}}
    {}^G\mathbf{n}) + \mathbf{n}_m,
    \end{aligned}
\end{equation}
where $\mathbf{n}_m$ is the plane measurement noise, whose covariance is given by \cite{yang2019tightly}.
We linearize this equation and obtain the following residual and Jacabians:
\begin{equation}
\begin{aligned}
\tilde{\mathbf{\Pi}}_m &=
\mathbf{H}_{T_{\Pi}}
\tilde{\mathbf{x}}_{T_m}
+ 
\mathbf{H}_{f_{\Pi}}
{}^G\tilde{\mathbf{\Pi}}
+ \mathbf{n}_{m}
,
\end{aligned}
\end{equation}
where $\tilde{\mathbf{\Pi}}_m$ is the 3D plane measurement residual, 
$\mathbf{H}_{T_{\Pi}}$ and $\mathbf{H}_{f_{\Pi}}$ are the measurement Jacobians in respect to the current state $\mathbf{x}_{T_m}$ and the plane feature ${}^G\mathbf{\Pi}$, respectively.

With initial estimations and an adequate amount of measurements, we perform an EKF update for plane features.
Normally, planes are tracked long but discontinuously between frames due to the instability of detecting them from noisy depth maps.
Thus, for planes, it is unreasonable to classify the long-term and short-term features based on the number of consecutive tracks.
Different from point features, all the planes are regarded as SLAM features and added to the state vector as soon as there are sufficient observations.
Moreover, to improve computational efficiency, small and short-tracked planes are delayed marginalized from the state when they are lost for an extended period (greater than 200 frames in our experiments).
Other planes will be maintained in the state as permanent landmarks providing long-term constraints.
In this way, we retain reliable and dominant planes in the state for overcoming cumulative drift.


\subsection{Drift Suppression}\label{secF}

To suppress potential drift, we introduce a novel graph-based strategy that fully exploits the spatial relations of plane patches to detect drift, followed by a de-drift update to correct the error.
More concretely, planes are organized as a graph that encodes geometric information of the scene. By matching two graphs, we search overlapping and identical structures in the global map, which are assumed to be inconsistencies caused by drift. The error is then corrected by aligning the detected structural `ghosting'.
In what follows, we will elaborate on the process.

\subsubsection{Problem Formulation}
Considering two graphs $\mathcal{G}_A = (\mathcal{V}_A, \mathcal{E}_A)$ and $\mathcal{G}_M = (\mathcal{V}_M, \mathcal{E}_M)$, the problem of graph matching can be formulated as determining an assign matrix $\mathbf{X}^*$:
\begin{equation}\label{obj}
\begin{aligned}
&\mathbf{X}^* = \arg \max
\sum_{i \in \mathcal{V}_A,j \in \mathcal{V}_M} \mathbf{X}_{ij}\mathbf{K}^P_{ij} + \\
&\sum_{(i_1,i_2) \in \mathcal{E}_A,(j_1,j_2)\in \mathcal{E}_M} \mathbf{X}_{i_1j_1}\mathbf{X}_{i_2j_2}\mathbf{K}^Q_{(i_1,i_2)(j_1,j_2)}
\\
s.t. \ & \mathbf{X}_{ij} \in \{0,\ 1\}, \ 
\mathbf{X}\mathbf{1}_{n_M} \le \mathbf{1}_{n_A},
\ 
\mathbf{X}^{\mathsf{T}} \mathbf{1}_{n_A} \le \mathbf{1}_{n_M},
\end{aligned}
\end{equation}
where $\mathcal{V}$ and $ \mathcal{E}$ stand for the vertex set and the edge set, and $n$ is the number of vertices.
$\mathbf{K}^P$ and $\mathbf{K}^Q$ are affinity matrices representing the similarity of vertices and edges, respectively.
Based on defined affinity matrices, the problem can be solved by factorized graph matching (FGM) \cite{zhou2015factorized}.

\subsubsection{Graph Construction}
In the context of graph matching, a plane $i$ is treated as a finite patch with several geometric attributes:
\begin{itemize}
    \item ${I}_i$: plane identity in the global map.
    \item $\mathbf{n}_i$: plane normal vector.
    \item $d_i$: distance from the origin to the plane.
    \item $\mathcal{L}_i$: list of convex hull points, which is computed in the plane detection thread after projecting all the fitted inliers onto the plane.
    \item $a_i$: area of the convex hull.
\end{itemize}

And four attributes are defined to describe the relation between two plane patches $(i,j)$: 
\begin{itemize}
    \item $\theta_{ij}$: angle of their normal vectors, $\theta_{ij} = \arccos (|\mathbf{n}_i^{\mathsf{T}} \mathbf{n}_j|), \ 0 \le \theta_{ij} \le \frac{\pi}{2}$.
    
    \item $d_{ij}$: minimum distance from all points on patch $i$ to patch $j$.
    
    \item $c_{ij}$: category of the relation. There are three types: a) If $\theta_{ij} \le \delta_{\theta}$, the category is `parallel'.
    b) Else if patch $j$ is separated from patch $i$ ($d_{ji} > 0$), the category is `separation'.
    c) Otherwise, the category is `intersection' that means patch $j$ is split by infinite plane $i$ ($d_{ji} = 0$).
    
    \item $o_{ij}$: overlapping area after projecting patch $i$ onto patch $j$ ($o_{ij} \ge 0$). When they are not overlapping, $-o_{ij}$ indicates the minimum parallel distance along patch $j$ between their convex hull points ($o_{ij} < 0$).
\end{itemize}

Note that the relations are asymmetrical because $c_{ij} \ne c_{ji}$, $d_{ij} \ne d_{ji}$, and $o_{ij} \ne o_{ji}$.
Fig. \ref{fig:relation} exhibits possible relative spatial positions of two plane patches.

\begin{figure}[b]
  \centering
\includegraphics[width=0.47\textwidth]{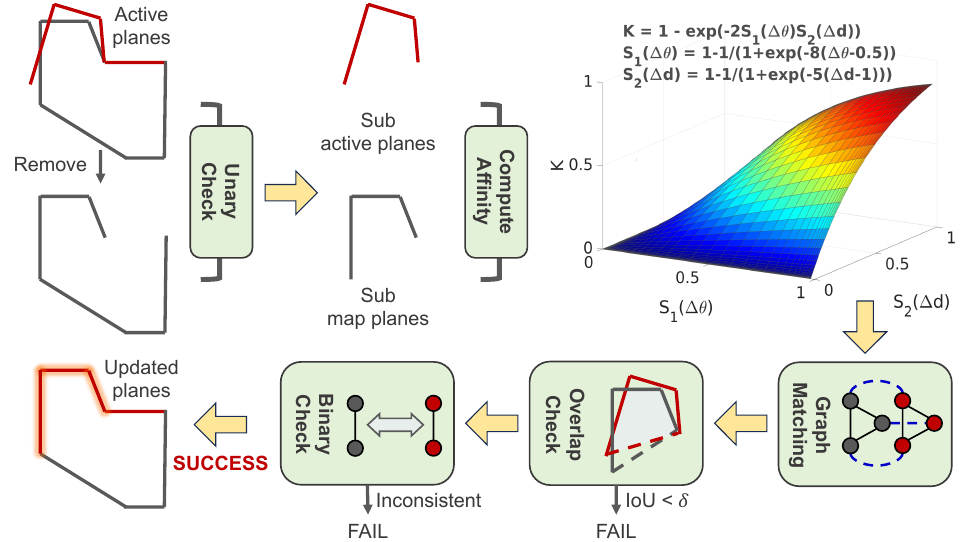}
  \caption{Pipeline of the drift suppression. 
  }
  \label{fig:graph}
\end{figure}

Based on these attributes, a scene can be represented as a directed graph, whose vertices and edges are plane patches and their geometric relationships. 
The angle and the distance are used to define the aforementioned affinity matrices and other attributes serve for validation. 
In particular, for vertices, the affinity matrix is defined as:
\begin{equation}
\begin{aligned}
\mathbf{K}^P_{ij} &= \mathcal{S}(\theta_{ij}, d_{ij}),
\end{aligned}
\end{equation}
and for edges it is:
\begin{equation}
\begin{aligned}
\mathbf{K}^Q_{({i_1},{i_2})({j_1},{j_2})} &= \mathcal{S}(\Delta \theta, \Delta d) \\
\Delta \theta =  |\theta_{{i_1}{i_2}} - \theta_{{j_1}{j_2}}|, &\
\Delta d = |d_{{i_1}{i_2}} - d_{{j_1}{j_2}}|,
\end{aligned}
\end{equation}
where $\mathcal{S}(\cdot, \cdot)$ is a score function 
mapping variables to $[0,\ 1]$, as plotted in Fig. \ref{fig:graph}.

We emphasize that the suggested graph differs from others because instead of considering the size and the center distance as comparisons,
we model the planes using convex hulls and fully explore the relative positional relationships at the plane boundaries to form directed edges, under the influence of the partial observation problem in the incremental plane map that the sizes of planes continue to expand during observation.

\begin{algorithm}
\caption{Graph-Based Drift Detection}
\label{alg2}
\setstretch{0.9} 
\textbf{Unary Check:}
    \begin{itemize}
        \item Retrieve candidate matches as constraints $\mathbf{Ct}$:
        ($\mathbf{Ct}_{ij} = 1$ means that plane $i$ in the local map can be matched with plane $j$ in the global map, otherwise it cannot.)
        \begin{itemize}
            \item \textbf{If} $\theta_{ij} < \delta_{\theta}$, $d_{ij} < \delta_{d}$, and $o_{ij} < \delta_{o}$
            \textbf{then} $\mathbf{Ct}_{ij} = 1$.
            \item \textbf{Else} $\mathbf{Ct}_{ij} = 0$.
        \end{itemize}
        
    \item Set active planes to be unmatched as priors:
    \begin{itemize}
        \item 
        \textbf{If} $I_j == I_i$ 
        \textbf{then} $\mathbf{Ct}_{:j} = 0$.
    \end{itemize}
    \item Take all candidate planes that satisfy $\sum {\mathbf{Ct}}_{i:} > 0$ and $\sum {\mathbf{Ct}}_{:j} > 0$ to build two subgraphs ${}^G\mathcal{G}_A'$ and ${}^G\mathcal{G}_M'$ with subconstraints $\mathbf{Ct}'$.    
\end{itemize}
\textbf{Graph Matching:}
\begin{itemize}
    \item Compute affinity matrices $\mathbf{K}^P$ and $\mathbf{K}^Q$ for ${}^G\mathcal{G}_A'$ and ${}^G\mathcal{G}_M'$.
    \item Get matches using the FGM algorithm:
        \begin{itemize}
        \item \textbf{Input: }
    ${}^G\mathcal{G}_A'$,
    ${}^G\mathcal{G}_M'$, $\mathbf{Ct}'$, 
        $\mathbf{K}^P$,
        $\mathbf{K}^Q$
        \item \textbf{Output: } $\mathbf{X}'$
        \end{itemize}
    \item Convert 
    $\mathbf{X}'$ to the original assign matrix $\mathbf{X}$.
\end{itemize}
\textbf{Overlap Check:}
\begin{itemize}
\item Compute the overall overlap of the matched vertices:
\begin{itemize}
\item Project the matched planes onto the ground (with known gravity) and calculate the overlap between convex hulls of each set, as illustrated in Fig. \ref{fig:graph}.
\end{itemize}
\item \textbf{If} $ o_{\mathcal{A}\mathcal{M}} / (a_{\mathcal{A}} + a_{\mathcal{M}} -o_{\mathcal{A}\mathcal{M}}) < \delta_{o}''$ \textbf{then} reject the matches.
\end{itemize}

\textbf{Binary Check:}

\begin{itemize}

\item Compare edge pairs for the matches:
\begin{itemize}
\item $ \Delta \theta =  |\theta_{{i_1}{i_2}} - \theta_{{j_1}{j_2}}| $
\item 
$ \Delta d = |d_{{i_1}{i_2}} - d_{{j_1}{j_2}}|$
\item
$ \Delta o = |o_{{i_1}{i_2}} - o_{{j_1}{j_2}}| $
\end{itemize}

\item \textbf{If}
 $ \Delta \theta  < \delta_{\theta}'$,
$ \Delta d  < \delta_{d}'$, 
and
$ \Delta o  < \delta_{o}'$ 
\textbf{then} accept it.
\item \textbf{Else} $\mathbf{X}_{i_1j_1} = 0, \mathbf{X}_{i_2j_2} = 0$ and reject the two matches.
\end{itemize}
\end{algorithm}

\subsubsection{Drift Detection}

The objective of drift detection is to search for similar and overlapping plane configurations from the global map, which is addressed as a problem of graph matching.
Algorithm \ref{alg2}  and Fig. \ref{fig:graph} outline the process.
Firstly, planes observed in the latest ten frames are considered as currently active planes (local map), and their nearest observations are constructed into a fully-connected graph.
Then we remove them from the global map and match them with the rest planes leveraging the FGM algorithm with the above defined affinity matrices.
In order to improve efficiency and robustness, we perform a unary check on the vertices to 
limit the number of candidate matches.
The thresholds here are relatively weak constraints to avoid rejecting correct matches under large drift.
In addition, since FGM may encounter failure modes, we adopt an overlap metric to quantify the overlap degree between the two configurations and employ a binary check to validate if the matched edges have similar relative spatial positions.
There may be occlusions between planes due to changes in viewpoints.
Therefore, instead of requiring the edges to be of the same category (as in Fig. \ref{fig:relation}), we only constrain their relative positions $\Delta \theta$, $\Delta d$, and $\Delta o$ within strict thresholds.
If all the constraints are satisfied, we accept the matches and consider that drift happens as the two graphs encode the same information.
The drift detection strategy requires at least three planes in the local map.

%

\subsubsection{De-Drift Update}
Once drift is detected, we fix the previously created landmark $\mathbf{\Pi}_r$ that is considered drift-free and enforce a pair-wise equality constraint to update the drifting plane landmark $\mathbf{\Pi}_d$:
\begin{equation}
\mathbf{z}_r = {}^G{\mathbf{\Pi}}_{d} - {}^G{\mathbf{\Pi}}_{r} + \mathbf{n}_r, 
\end{equation}
where $\mathbf{z}_r$ is the drift residual and $\mathbf{n}_r$ is a random noise.
Following that, the system state is refined with fixed plane landmarks.
Eventually, similar planes will be merged in delayed marginalization.


\begin{table*}[thpb]
\centering
\caption{Evaluation on the CID-SIMS Dataset (RMSE ATE ($\downarrow$) in Meters)}
\label{table:cid}
\resizebox{1.85\columnwidth}{!}
{
\begin{threeparttable}
\begin{tabular}{cc|ccccc|ccc}
\toprule[1.5pt]
Sequence   & Length {[}m{]} & ORB-SLAM3 \cite{2021orbslam3} & VINS-Mono \cite{qin2018vins} & OpenVINS  \cite{geneva2020openvins} & ov\_plane \cite{chen2023monocular} & PlanarSLAM \cite{li2021rgb} & PGD w/o P. & PGD w/o G. & PGD-VIO \\
\bottomrule
Office\_1     & 42.01          & 0.104         & 0.120     & 0.155    & 0.133          & 0.228      & \underline{0.097}          & \textbf{0.030} & \textbf{0.030} \\
Office\_2     & 90.31          & 0.496          & 0.158    & 0.079    & 0.102          & 0.275      & 0.091          & \underline{0.030} & \textbf{0.029} \\
Office\_3     & 95.25          & 0.062          & /    & 0.112    & 0.110          & 0.311      & 0.123          & \textbf{0.033} & \underline{0.039} \\
Floor14\_1    & 103.7          & 0.260          & 0.412         & 0.728    & 0.408          & 0.265      & 0.229          & \underline{0.173}          & \textbf{0.132} \\
Floor14\_2    & 106.4          & 3.569          & 2.627         & 1.750    & 1.704          & /          & \underline{0.325} & 0.367          & \textbf{0.163} \\
Floor14\_3    & 180.43         & 0.415           & /     & 0.494    &\textbf{0.345} & 5.303      & \underline{0.382}          & 0.399          & 0.645          \\
14-13-14      & 249.96         & \textbf{0.416} & /         & 1.510    & 1.518          & /          & 0.992          & 0.861          & \underline{0.824} \\
14-13-12      & 21.89          & 3.042              & 0.773     & 0.118    & \textbf{0.106} & 1.878      & 0.119          & \underline{0.108} & \underline{0.108} \\
Floor3\_1     & 85.61          & \underline{0.283}  & /     & 0.833    & 0.655          & 1.037      & 0.547          & 0.476          & \textbf{0.112} \\
Floor3\_2     & 150.55         & 4.877          & 0.525     & 1.336    & 1.591          & 0.886      & 0.799          & \underline{0.426} & \textbf{0.396} \\
Floor3\_3     & 196.46         & \textbf{0.323} & 2.073     & 2.127    & 3.026          & /          & 0.733          & 1.372          & \underline{0.692} \\
Floor13\_1    & 130.43         & 0.921          & /     & /        & 1.721          & /          & 0.912          & \underline{0.434}          & \textbf{0.380} \\
Floor13\_2    & 135.10         & \underline{0.494} & 2.655     & 1.415    & /              & /          & 2.009          & 0.498          & \textbf{0.403} \\
\midrule
Apartment1\_1 & 66.01          & 0.187     & 0.397     & /        & 0.476          & 0.135      & 0.139          & \underline{0.050}          & \textbf{0.049} \\
Apartment1\_2 & 77.18          & /              & /         & 0.185    & \textbf{0.074} & 0.212      & 0.129          & \underline{0.085} & 0.136          \\
Apartment1\_3 & 154.00         & 0.688      & 0.355     & 0.224    & 0.261          & /          & 0.234          & \underline{0.163}          & \textbf{0.138} \\
Apartment2\_1 & 68.50          & 2.221          & 0.216     & /        & /              & 0.199      & 0.114          & \textbf{0.045} & \underline{0.050} \\
Apartment2\_2 & 85.88          & 0.739              & 0.146     & 0.181    & 0.272          & /          & 0.107          & \underline{0.053} & \textbf{0.051} \\
Apartment2\_3 & 100.04         & 0.096       & 0.252         & 0.144    & /              & /          & 0.104          & \underline{0.049}          & \textbf{0.041} \\
Apartment3\_1 & 73.22          & 0.111       & 0.508     & 0.122    & 0.145          & 0.747      & 0.115          & \underline{0.076}          & \textbf{0.065} \\
Apartment3\_2 & 84.42          & 0.292       & /         & 1.068    & 0.097          & /          & \textbf{0.076} & \underline{0.090}          & 0.097          \\
Apartment3\_3 & 147.96         & 2.689         & 0.346         & 0.144    & 0.137          & 0.174      & 0.119          & \textbf{0.074} & \underline{0.089}\\  
\bottomrule[1.5pt]
\end{tabular}
\begin{tablenotes}
 \item * The best results for each sequence are boldfaced and the next best results are underlined.
\end{tablenotes}
\end{threeparttable}
}
\end{table*}

\begin{figure*}[thpb]
  \centering
\includegraphics[width=0.43\textwidth]{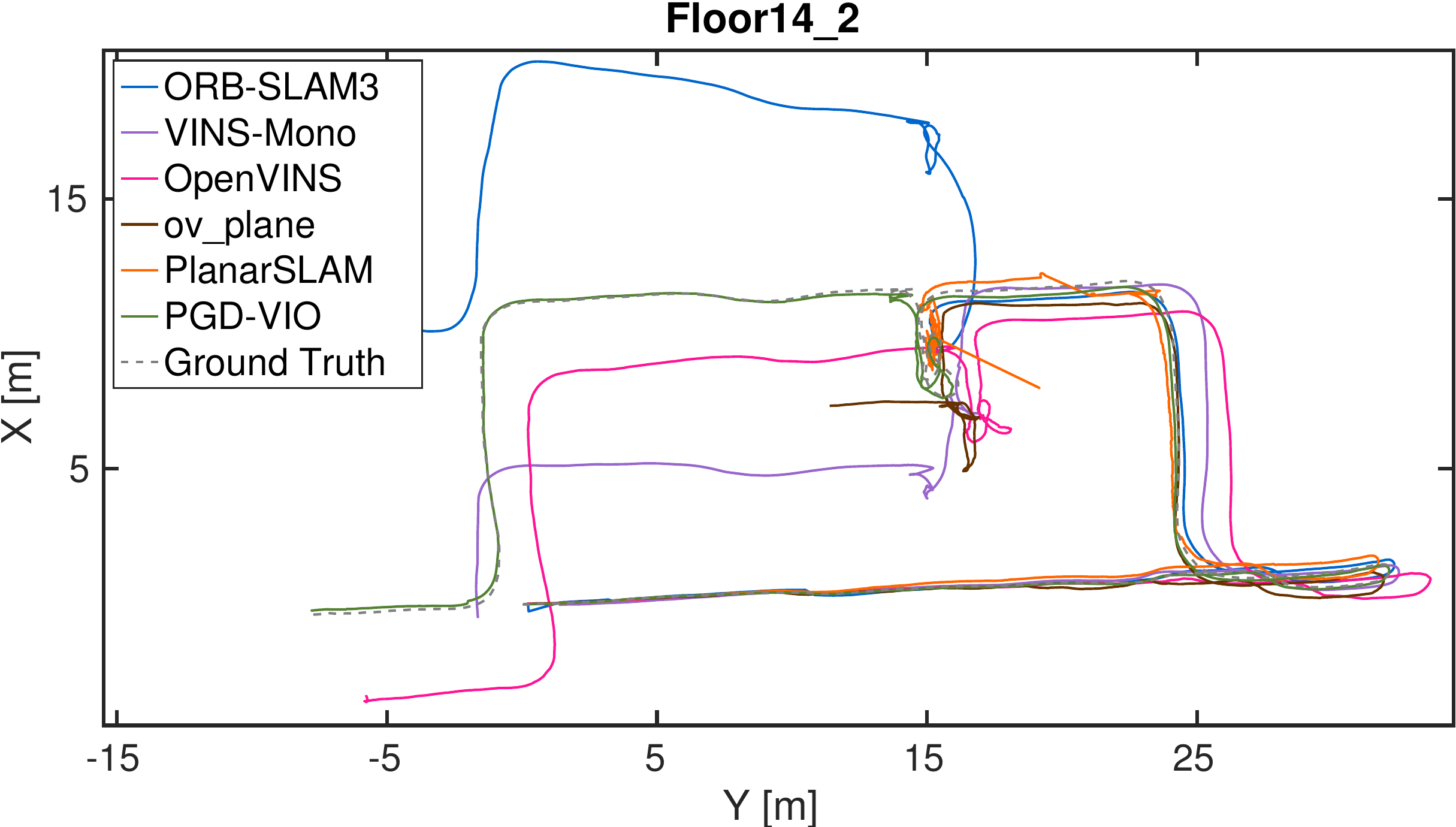}
\includegraphics[width=0.229\textwidth]{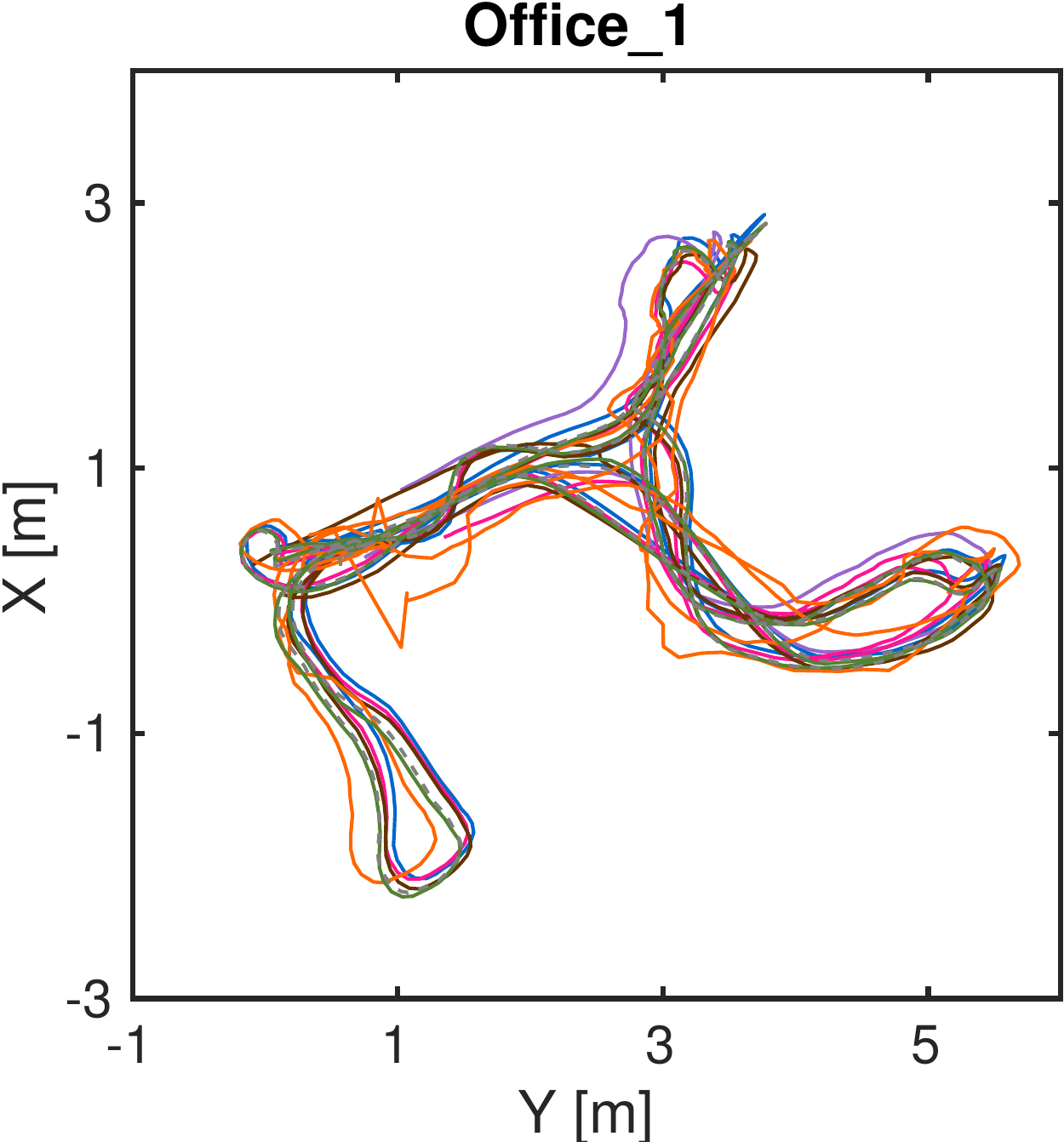}
\includegraphics[width=0.23\textwidth]{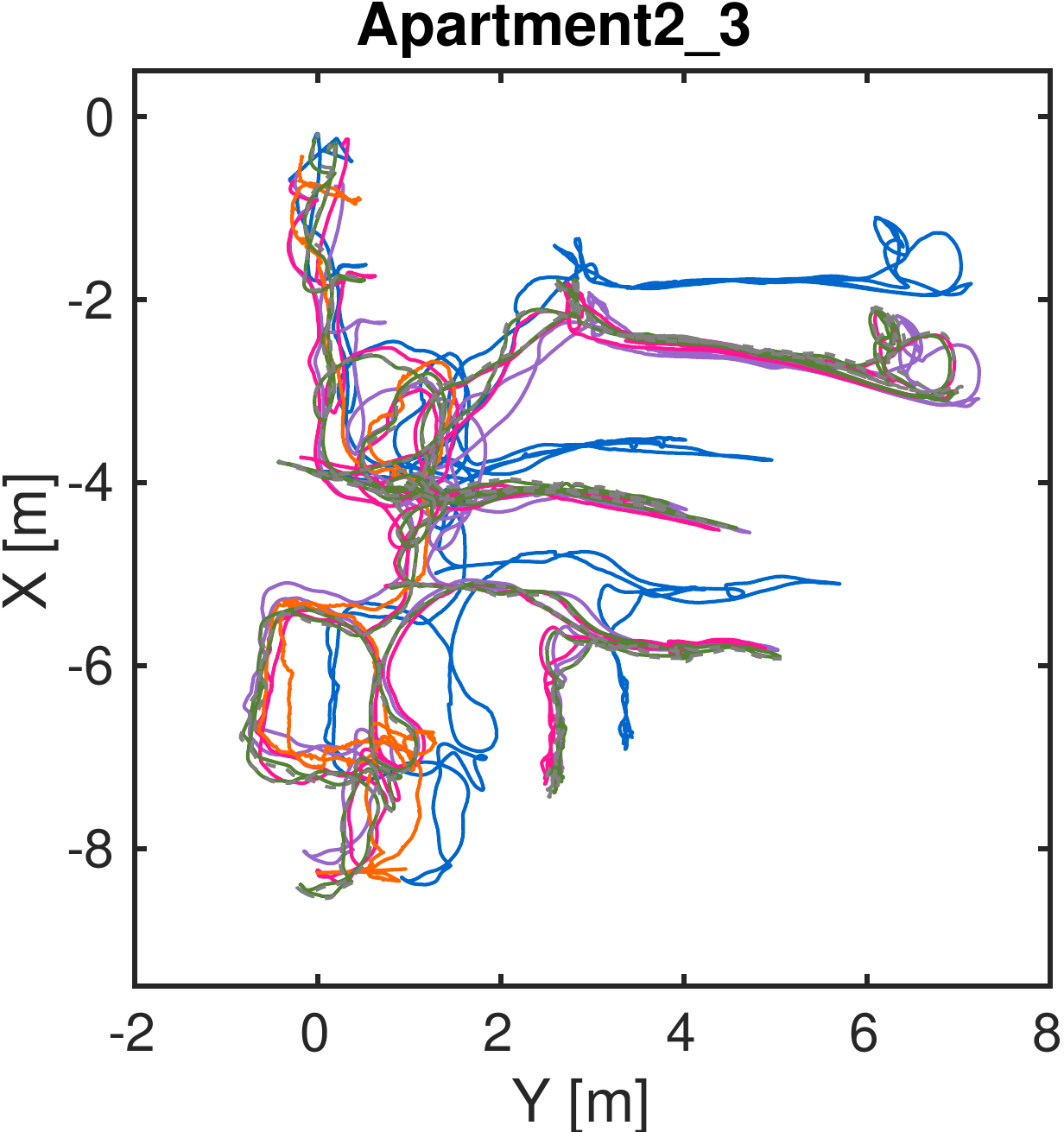}
  \caption{Comparative trajectories of the evaluated methods on the CID-SIMS dataset. For visualization, the first 500 frames are used to align the trajectories with the ground truth.}
  \label{fig:cidalltrajs}
\end{figure*}

\section{EXPERIMENTS}
In this section, we evaluate the overall performance of the proposed PGD-VIO on two public RGB-D inertial datasets: the CID-SIMS dataset \cite{zhang2023cid} and the VCU-RVI dataset \cite{zhang2020vcu}.

We make comparisons with well-known point-based systems (ORB-SLAM3 \cite{2021orbslam3}, VINS-Mono \cite{qin2018vins}, and OpenVINS \cite{geneva2020openvins}) and two plane-aided systems (ov\_plane \cite{chen2023monocular} and PlanarSLAM \cite{li2021rgb}).
All the experiments are performed on an Intel i9-13900KS CPU with suggested configurations.
The root mean square error (RMSE)
of the absolute trajectory error (ATE) is considered as the quantitative evaluation criterion.
We disable the global bundle adjustment module of ORB-SLAM3 and adopt the online poses for a fair comparison.
Additionally, considering the randomness of pose estimation, we run each system five times and report the median results.
/ indicates the method fails in all five tests when the estimated trajectory is less than 50\% complete or drifts larger than 10 m.
We also perform ablation studies considering two variants of our method.
\textbf{PGD w/o P.} is the RGB-D-inertial mode after introducing depth information.
On this basis,
\textbf{PGD w/o G.} adds plane measurements into the state vector to provide a baseline for \textbf{PGD-VIO}, which is the full version with the proposed graph-based drift suppression.

\subsection{CID-SIMS Dataset}
The CID-SIMS dataset \cite{zhang2023cid} is a challenging indoor dataset for wheeled robots with abundant real environments and provides the whole ground truth for long sequences.
According to the results from Table \ref{table:cid},
PGD-VIO achieves the lowest or the second-lowest ATE in most sequences, free of expensive global bundle adjustment (BA) and loop closing techniques, exhibiting superior performance
to ORB-SLAM3, which
performs poorly in several challenging sequences because the tracking is lost in weakly textured regions and fast motions.
PlanarSLAM is an RGB-D system built on ORB-SLAM2 \cite{2017orbslam2} that estimates rotations based on the Manhattan structure assumption and optimizes translations in the BA.
Suffering from the same issues with ORB-SLAM3, PlanarSLAM easily collapses.
Specially, without the assistance of IMU measurements, PlanarSLAM fails in more sequences than other methods.
In long-term sequences, the degenerate movement of a wheeled robot, such as moving along straight lines, makes VINS-Mono and OpenVINS fail to observe the scale information, which brings about large locating errors.
ov\_plane extends OpenVINS by adding planes, in which points and planes are treated as combinations of SLAM features and MSCKF features for different updates.
Although it surpasses OpenVINS on average owing to the exploitation of point-on-plane constraints, the improvement is limited as few planes are successfully tracked in the clustered environments in this dataset.
Exemplary trajectories estimated by these methods and the corresponding ground truth are displayed in Fig. \ref{fig:cidalltrajs}. 
Overall, PGD-VIO has a tendency to show more complete trajectories with low localization errors compared to the other algorithms, benefiting from the proposed novelties.

\begin{figure*}[thpb]
  \centering
  \includegraphics[width=0.55\textwidth]{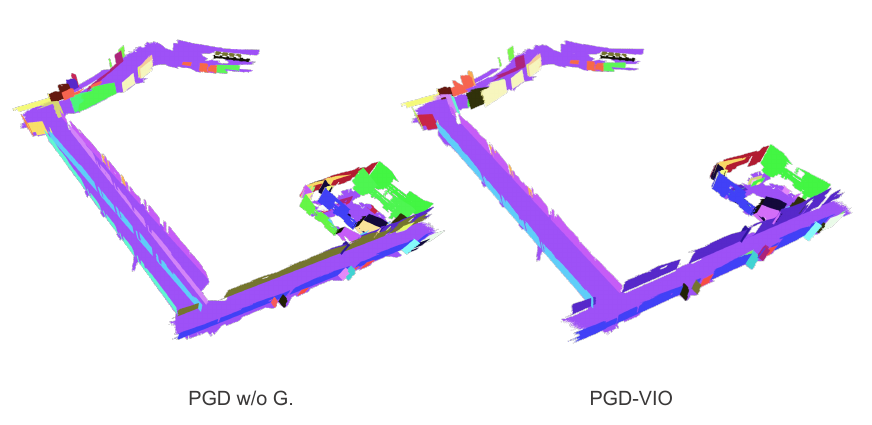}
  \includegraphics[width=0.29\textwidth]{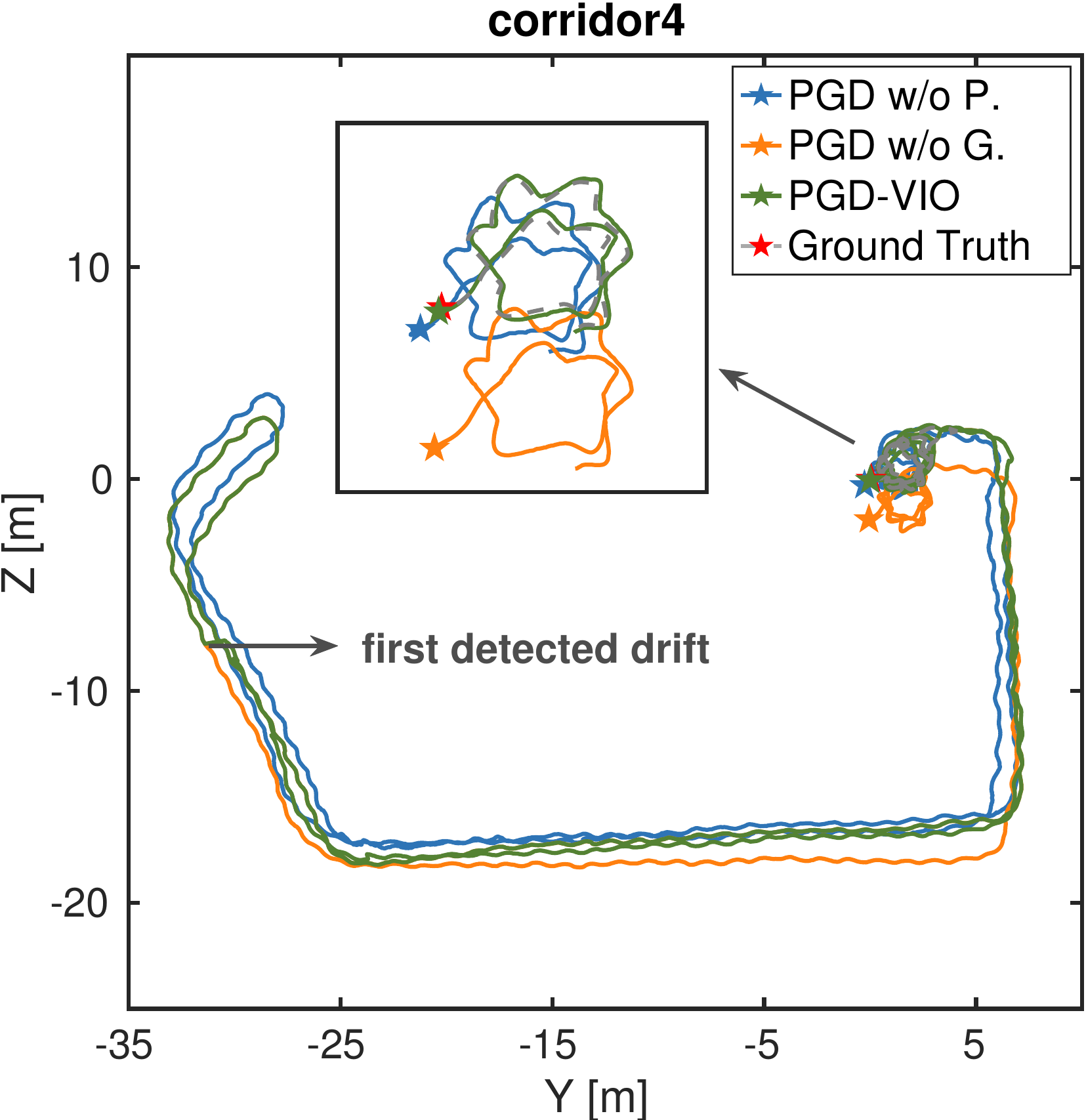}
  \caption{Results on sequence \textit{corridor4} (210 m) of the VCU\_RVI dataset. 
  The left two figures show the plane maps without and with drift suppression, respectively, and planes are visualized in different colors.
  The right figure displays the comparison trajectories, in which
  the first 500 frames are used to align the estimated trajectories with the ground truth trajectories and the end points marked as stars reflect their cumulative drift. 
  Before the first drift is detected, PGD w/o G. and PGD-VIO are completely coincident.
  The error of PGD w/o G. gradually increases over time, whereas PGD-VIO effectively reduces the final cumulative error and enhances the consistency of the final map with the help of drift suppression.}
\label{fig:vcu}
\end{figure*}

Regarding the influence of different modules, 
the absolute scale is supplemented after integrating depth information and thus PGD w/o P. has reasonable performance in all sequences, better than VINS-Mono and OpenVINS.
Compared with it, deploying plane landmarks for state updating brings an improvement for PGD w/o G. in 82\% sequences.
The performance degrades in some sequences, for example, sequence \textit{Floor14\_3} and \textit{Floor3\_3}, as it detects inaccurate plane measurements because of the reflective mirrors throughout the long corridors.
For short-term sequences that are full of rotations, e.g. sequences in office and apartment environments, the proposed drift suppression strategy slightly affects the system as no substantial drift occurs.
In contrast, PGD-VIO exhibits obvious advantages in long-term sequences, e.g. sequences in floor environments, by aligning the structural `ghostings’ in the scenes, which effectively proves that the drift suppression strategy is helpful to ease cumulative errors.



\begin{table}[b]
\centering
\caption{Evaluation on the VCU-RVI Dataset (RMSE ATE ($\downarrow$) in Meters)}
\label{table:vcu}
\resizebox{0.9\columnwidth}{!}{
\begin{threeparttable}
\begin{tabular}{ccccc}
    \toprule[1.5pt]
    Method & corridor1 & corridor2 & corridor3 & corridor4\\
    \midrule
    ORB-SLAM3 \cite{2021orbslam3}  & 0.488 & / & 5.652 & 4.089 \\
VINS-Mono \cite{qin2018vins}  & 4.390          & 1.610          & 3.970          & 4.330          \\
VINS-RGBD \cite{shan2019rgbd}  & 5.130          & 1.810          & 6.810          & 1.950          \\
OpenVINS  \cite{geneva2020openvins}   & 1.039          & 1.639          & 0.435          & 0.706          \\
ov\_plane \cite{chen2023monocular} & 1.298          & 3.226          & 0.922          & 1.262          \\
PlanarSLAM \cite{li2021rgb} & \textbf{0.073} & /              & /              & /              \\
S-VIO \cite{gu2023s}     & 0.580          & 1.490          & 0.910          & \underline{0.200}          \\
\midrule
PGD w/o P.  &  1.411  & 1.423 & 0.809 & 0.379 \\
PGD w/o G.  & 0.223          & \underline{0.285} & \underline{0.418} & 0.945 \\
PGD-VIO   & \underline{0.192}          & \textbf{0.075} & \textbf{0.404} & \textbf{0.121}\\
\bottomrule[1.5pt]
\end{tabular}
\begin{tablenotes}
 \item * The best results for each sequence are boldfaced and the next best results are underlined.
\end{tablenotes}
\end{threeparttable}
}
\end{table}

\subsection{VCU-RVI Dataset}
Furthermore, pose estimation accuracy and map consistency are evaluated on the VCU-RVI dataset \cite{zhang2020vcu}, which provides ground truth trajectories at the beginning and the ending to manifest cumulative drift for long-term sequences.
VINS-RGBD \cite{shan2019rgbd} and S-VIO \cite{gu2023s} are further included in the comparison using results from the original papers \cite{zhang2020vcu, gu2023s}.
We list the RMSE ATE for the corridor sequences in Table \ref{table:vcu}.
As evident, 
PGD-VIO achieves the lowest ATE except on sequence \textit{corridor1}, where PlanarSLAM performs well. However, PlanarSLAM fails in other sequences because point feature tracking is lost under conditions of insufficient textures and imperfect Manhattan structures.
Also, we observe that ov\_plane struggles with detecting planes in most images and therefore does not effectively improve the accuracy of OpenVINS.
We hope that plane measurements in PGD w/o G. contribute to localization as in sequence \textit{corridor1}, \textit{corridor2}, and \textit{corridor3}.
However, if the system experiences drift, there is a certain probability of erroneously associating unrelated planes, so that the system state is updated towards the wrong direction, and ultimately resulting in low accuracy, as in sequence \textit{corridor4}.
The proposed drift suppression strategy helps address the problem as early as possible to avoid large drift.
By virtue of it, PGD-VIO obtains a 45\% improvement on average compared to PGD w/o G. in these long corridor sequences.
Intuitively, Fig. \ref{fig:vcu} illustrates the effectiveness of the drift suppression module.


\section{CONCLUSIONS}
In this paper, we propose an RGB-D VIO system, named PGD-VIO, effectively integrating depth information and plane measurements within the naive EKF framework.
More importantly, we fully exploit different spatial relations of boundary plane patches and apply a graph-based strategy for drift suppression.
The proposed system is assessed on two real-world datasets with experimental results proving that PGD-VIO greatly enhances the performance against cumulative drift, enables robust and accurate localization without loop closures and produces highly consistent plane maps, especially in long-term navigation.
However, PGD-VIO struggles when planar structures are few or indistinguishable 
due to the repetitiveness of the scenes.
In the future, we aim to take into account the structure regularities in the process of plane association and EKF update to better explore the available geometrical information in planar environments.

\addtolength{\textheight}{-8cm}   



\bibliographystyle{IEEEtranS}
\bibliography{root.bib}

\end{document}